\documentclass{article}

\usepackage{microtype}
\usepackage{graphicx}
\usepackage{subfigure}
\usepackage{booktabs} % for professional tables
\usepackage{tabularx}

\usepackage{hyperref}

\usepackage[accepted]{icml2019}

\icmltitlerunning{Submission and Formatting Instructions for ICML 2019}

\begin{document}

\twocolumn[
\icmltitle{Triplet Distillation for Deep Face Recognition}

% It is OKAY to include author information, even for blind
% submissions: the style file will automatically remove it for you
% unless you've provided the [accepted] option to the icml2019
% package.

% List of affiliations: The first argument should be a (short)
% identifier you will use later to specify author affiliations
% Academic affiliations should list Department, University, City, Region, Country
% Industry affiliations should list Company, City, Region, Country

% You can specify symbols, otherwise they are numbered in order.
% Ideally, you should not use this facility. Affiliations will be numbered
% in order of appearance and this is the preferred way.
\icmlsetsymbol{equal}{*}

\begin{icmlauthorlist}
\icmlauthor{Yushu Feng}{to}
\icmlauthor{Huan Wang}{to}
\icmlauthor{Daniel T.~Yi}{goo}
\icmlauthor{Roland Hu}{to}
\end{icmlauthorlist}

\icmlaffiliation{to}{College of Information Science and Electronic Engineering, Zhejiang University, Zhejiang, China}
\icmlaffiliation{goo}{Gilman School, Baltimore, Maryland, USA}

\icmlcorrespondingauthor{Yushu Feng}{fengyushu@zju.edu.cn}

% You may provide any keywords that you
% find helpful for describing your paper; these are used to populate
% the "keywords" metadata in the PDF but will not be shown in the document
\icmlkeywords{Machine Learning, ICML}

\vskip 0.3in
]

% this must go after the closing bracket ] following \twocolumn[ ...

% This command actually creates the footnote in the first column
% listing the affiliations and the copyright notice.
% The command takes one argument, which is text to display at the start of the footnote.
% The \icmlEqualContribution command is standard text for equal contribution.
% Remove it (just {}) if you do not need this facility.

\printAffiliationsAndNotice{}  % leave blank if no need to mention equal contribution
%\printAffiliationsAndNotice{\icmlEqualContribution} % otherwise use the standard text.

\begin{abstract}
Convolutional neural networks (CNNs) have achieved a great success in face recognition, which unfortunately comes at the cost of massive computation and storage consumption.
Many compact face recognition networks are thus proposed to resolve this problem.
Triplet loss~\cite{schroff2015facenet} is effective to further improve the performance of those compact models.
However, it normally employs a fixed margin to all the samples, which neglects the informative similarity structures between different identities.
In this paper, we propose an enhanced version of triplet loss, named \emph{triplet distillation}, which exploits the capability of a teacher model to transfer the similarity information to a small model by adaptively varying the margin between positive and negative pairs.  
Experiments on LFW, AgeDB, and CPLFW datasets show the merits of our method compared to the original triplet loss.
\end{abstract}
% ------------------------------------------------------------------------------

\section{Introduction}
\label{intro}

Recent years have witnessed the impressive success of CNNs in the area of face recognition~\cite{parkhi2015deep,sun2014deep1,taigman2014deepface,sun2014deep2}.
However, effective face recognition CNN models typically consume a large amount of storage and computation, making it difficult to deploy on mobile and embedded devices.
To resolve this problem, several lightweight CNN models have been proposed, such as MobileID~\cite{luo2016face}, ShiftFaceNet~\cite{wu2018shift}, and MobileFaceNet~\cite{chen2018mobilefacenets}.
Unfortunately, model size reduction usually coincides with performance decline.
Triplet loss~\cite{schroff2015facenet}, as a metric learning method, is widely used in face recognition to further improve accuracy~\cite{deng2018arcface}.
Triplet loss explicitly maximizes the inter-class distance and meanwhile minimizes the intra-class distance, where a margin term is used to determine the decision boundaries between positive and negative pairs.

In the original triplet loss, the margin is set to a constant, which tends to push the decision boundaries among different classes to the same value, thus loses the hidden similarity structures of different identities.
Therefore, it is necessary to set a~\emph{dynamic} margin to take into account the similarity structures.
In this vein, \cite{zakharov20173d} sets the margin term as a function of angular differences between the poses for pose estimation;~\cite{wang2018deeprank} formulates the adaptive margin as a nonlinear mapping of the average distances among different people for person re-identification. 
However, they obtain the dynamic margins by handcrafted rules rather than learned distances.
In this paper, we propose an enhanced version of triplet loss, named \emph{triplet distillation}, which borrows the idea of knowledge distillation~\cite{hinton2015distilling} to determine the dynamic margins for face recognition.
Specifically, we determine the similarity between two identities according to distances learned by the teacher model.
This similarity, as a kind of knowledge, is then applied to guiding the student model to optimize its decision boundaries.

The major contributions of this work can be summarized as follows:
\begin{itemize}
\itemsep=-2pt
\item We propose the triplet distillation method to transfer knowledge from a teacher model to a student model for face recognition.  
\item We improve the triplet loss with dynamic margins by utilizing the similarity structures among different identities, which is in contrast with the fixed margin of the original triplet loss.
\item Experiments on LFW~\cite{huang2008labeled}, AgeDB~\cite{moschoglou2017agedb} and CPLFW~\cite{CPLFWTech} show that the proposed mehtod performs favorably against the original scheme.
\end{itemize}
% ---------------------------------------------------------------------

\section{Related Work}
\label{related}
\textbf{Triplet loss.} The main purpose of triplet loss~\cite{schroff2015facenet} is to distinguish identities in the projected space with the guidance of distances among an anchor sample, a positive sample, and a negative sample. 
There are several revisions for the original triplet loss, which mainly fall into the following three categories:
(1) Adding new constraints to the objective function to improve the generalization performance~\cite{cheng2016person,chen2017beyond};
(2) Optimizing the selection of triplet samples to make the triplet samples more informative, which can lead to faster convergence and better performance~\cite{sohn2016improved,hermans2017defense,ge2018deep,dong2018triplet,ming2017simple};
(3) Proposing dynamic margins for different triplet combinations, such as \cite{zakharov20173d, wang2018deeprank} which use handcrafted methods to determine the similarities among different identities.
Our method belongs to the last category. Different from previous approaches, we exploits a teacher model to obtain the similarity information among identities to set the dynamic margins.

\textbf{Knowledge distillation.} 
Knowledge distillation, firstly proposed by \cite{buciluǎ2006model} and then refined by Hinton et al.~\cite{hinton2015distilling}, is a model compression method to transfer the knowledge of a large teacher network to a small student network.
The main idea is to let the student network learn a mapping function which is similar to the teacher network.
Most researches follow~\cite{hinton2015distilling} to learn the soft-target outputs of the teacher network~\cite{fukuda2017efficient, sau2016deep, zhou2018rocket, furlanello2018born}. 
These methods make the student model match the output distributions of the teacher model. 
Not confined to the output distributions of the teacher model, the definition of knowledge can also refer to its feature maps. For example, \cite{romero2014fitnets, huang2017like, chen2018learning} utilize feature maps of the middle layers to guide the knowledge transfer from the teacher model to the student model. 
Recent works further broaden the definition of knowledge to other attributes such as  attention maps~\cite{zagoruyko2016paying, huang2017like} and affinity among training samples~\cite{chen2018darkrank}. 
In this paper, we also use the knowledge of feature similarity between identities as a guidance to train the student model.   

%------------------------------------------------------------------------------

\section{The Proposed Method}
\label{appro}
\subsection{Teacher and student networks}
\label{student network}
We employ the widely-used ResNet-100~\cite{he2016deep} as the teacher model.
For the student model, we adopt a slim version of MobileFaceNet~\cite{chen2018mobilefacenets}, which has the same architecture as MobileFaceNet, yet with three quarters of the number of channels in each convolutional layer on average.
The detailed statistics of the teacher and student model are summarized in Table~\ref{compa}.
%
% The main building block of the student model is the residual bottlenecks proposed in MobileNet-V2~\cite{sandler2018mobilenetv2}, each of which consists of a $1\times1$ convolution layer that expands the input channels with an expansion factor, a $3\times3$ depthwise convolution layer, and finally a $1\times1$ convolution layer to shrink the number of channels to its original size.
%
% MobileFaceNet has $5$ bottlenecks and the expansion factor in the first $1\times1$ convolution layer of each bottleneck is set to~$2$,$4$,$2$,$4$,$2$, respectively.
%
% In our smaller network, all of these factors are set to $4$.
%
% The output feature of our student model is 256-dimensions.

\begin{table}[h]
  \caption{Comparison between teacher model, MobileFaceNet~\cite{chen2018mobilefacenets}, and student model. The FLOPs are counted by TFProf, a profiling tool in Tensorflow. The inference time is averaged by $5000$ runs of forwarding an image of size $112\times112\times3$ on Intel Xeon(R) CPU E5-2609 v4 @1.70GHz with single thread.}
  \label{compa}
    \vskip 0.15in
    \centering
    \begin{small}
    \begin{tabular}{l p{0.9cm}<{\centering} p{1.2cm}<{\centering} p{1.2cm}<{\centering} p{0.8cm}<{\centering} }
        \toprule
        Model & Size/MB & Params/$10^6$ & FLOPs/$10^9$ & Time/s \\
        \midrule
        Teacher model   & $248.8$ & $652.25$ & $24.23$ & $2.45$ \\
        MobileFaceNet   & $4.0$ & $0.99$ & $0.49$ & $0.31$ \\
        Student model   & $2.3$ & $0.59$ & $0.23$ & $0.18$ \\
        \bottomrule
    \end{tabular}
    \end{small}
\end{table}

% --------------------------------------------------------------------------------------------

\subsection{Triplet distillation}
Triplet loss is applied to a triplet of samples, represented as~$\{x^a, x^p, x^n\}$. Here~$x^a$ is the anchor image; $x^p$ is called the positive image, which belongs to the same identity as~$x^a$, and~$x^n$ is called the negative image, which belongs to a different identity of~$x^a$. 
The triplet loss aims to minimize the distance between the anchor and positive images, and meanwhile maximize the distance between the anchor and negative images. 
The objective function of triplet loss can be formulated as
\begin{equation}
\label{eqn:triplet_loss}
\mathcal{L}=\frac{1}{N}\sum_{i}^{N}\max(\mathcal{D}(x_i^a,x_i^p)-\mathcal{D}(x_i^a,x_i^n)+m,0),
\end{equation}
where $N$ is the number of triplets in a mini-batch;~$\mathcal{D}(*,*)$ denotes the distance between two images.
Notably, the hyper-parameter~$m$ represents a margin enforced between the positive and negative pairs, that is, only when the distance difference between the negative pair and the positive pair is larger than a threshold~$m$, will the loss~$\mathcal{L}$ not count.
Naturally, the final distances among different identity clusters will be pushed to the margin $m$.
% so~$m$ can be interpreted as the expected minimal safe distance among different identity clusters.

%Triplet Loss can be shown in Figure.~\ref{Triplet Loss}.
%\begin{figure}[ht]
%\vskip 0.02in
%\begin{center}
%\centerline{\includegraphics[width=0.5\columnwidth]{figures/triplet_loss.png}}
%\caption{Triplet Loss. Anchor (a) and Positive (p) belong to the same identity. Negative(n) is from a %different identity. the Margin is enforced between positive and negative pairs. It takes the distance %between intra-class and inter-class as the optimization goal.}
%\label{Triplet Loss}
%\end{center}
%\vskip 0.02in
%\end{figure}

In the original triplet loss, $m$ is the \emph{same} for all identities. In other words, all identity clusters will be separated with a roughly same distance, which ignores the subtle similarity structures among different identities, since different people are not equally different.
For example, if person~$A$ looks more similar to person~$B$ than to person~$C$, then it should be better to set the~$m$ for \{$A$, $B$\} smaller than the~$m$ for \{$A$, $C$\} because such setting will push~$A$ and~$B$ closer than~$A$ and~$C$ in the hyperspace of the student model. 
In a similar spirit to dark knowledge proposed in knowledge distillation~\cite{hinton2015distilling}, this similarity structure is informative and useful, but not considered in the original triplet loss.
Our proposed triplet distillation method exploits knowledge distillation to bridge this gap.

First, the teacher model extracts two features from a triplet and obtains the distance between them. 
Then, we map this distance into the margin and apply it to the training of the student model.
Different from previous mathematical angle calculation methods~\cite{zakharov20173d,wang2018deeprank}, our scheme adopts the well-trained teacher model to calculate the face distance, which has more capability to capture the similarity structures in its learned representations.
With the proposed dynamic margin term, the objective function can be written as
\begin{equation}
\mathcal{L}=\frac{1}{N}\sum_{i}^{N}\max(\mathcal{D}(x_i^a,x_i^p)-\mathcal{D}(x_i^a,x_i^n)+\mathcal{F}(d),0),
\end{equation}
\begin{equation}
\label{eqn:d}
d=\max(\mathcal{T}(x_i^a,x_i^n)-\mathcal{T}(x_i^a,x_i^p),0), 
\end{equation}
where ~$\mathcal{D}(*,*)$ denotes the distance between two images calculated by the student model, ~$\mathcal{T}(*,*)$ represents the distance calculated by teacher model, $d$ denotes the distance between intra-class and inter-class features extracted by the teacher model, and $\mathcal{F}(*)$ represents the function of the margin with regards to the distance.
We employ a simple increasing linear function for $\mathcal{F}(*)$, 
\begin{equation}
\mathcal{F}{(d)}=\frac{m_{max}-m_{min}}{d_{max}}{d}+m_{min}
\end{equation}
where $m_{min}$ and $m_{max}$ represent the minimum and maximum values of margin; and~$d_{max}$ represents the maximum distance in a mini-batch.
In this way, the margin is constrained between $m_{min}$ and $m_{max}$.

\section{Experiments}
\label{exper}
\subsection{Implementation details}
\textbf{Pre-processing.} 
We use MTCNN~\cite{zhang2016joint} to detect faces and facial landmarks on the MS-Celeb-1M dataset~\cite{guo2016ms}, which consists of~$10$ million photos of~$100k$ celebrities.
To obtain data of higher quality, $3.8$ million photos from $85$ identities are picked out to make a refined MS-Celeb-1M dataset~\cite{deng2018arcface}.
% by ranking all face images of each identity based on their distances to the identity center
%
All the images are aligned based on the detected landmarks and then resized to $112\times112$ with normalization (subtracted by mean~$127.5$ and divided by standard deviation~$128$).

\textbf{Training.} The architectures of the teacher and student models are described in Section~\ref{student network}. 
Both of them are first trained from scratch with the ArcFace loss~\cite{deng2018arcface}.
Stochastic Gradient Descent (SGD) is used with momentum $0.9$ and batch size $480$.
The learning rate begins with $0.1$ and is divided by $10$ at iteration $30k$ and $70k$,  before the training finally ends at iteration $100k$.

Then the proposed triplet distillation is used to fine-tune the student model.
During this stage, there are~$10$ classes, $18$ images per class in each mini-batch.
The learning rate is $0.001$ and the training stops at $34k$ iterations.
We randomly sample $1000$ triplets from the refined MS-Celeb-1M dataset to obtain different $d$'s (Equation~(\ref{eqn:d})). Then the largest one is chosen as~$m_{max}=0.5$, and the smallest one as~$m_{min}=0.2$.
TensorFlow~\cite{abadi2016tensorflow} is used in all our experiments.
Our source codes and trained models will be made available to the public.

\textbf{Evaluation}.
In the evaluation stage, we extract the features of each image and its horizontally flipped image.
Then the two features are concatenated as one for face verification using the cosine distance.
Three popular face verification datasets are considered here: LFW~\cite{huang2008labeled}, CPLFW~\cite{CPLFWTech}, and AgeDB~\cite{moschoglou2017agedb}. 
For LFW and CPLFW, we adopt all the provided pairs ($3000$ positive and $3000$ negative pairs for each dataset); for AgeDB, which has $5$ different year gaps, we only choose one of them with $300$ positive and $300$ negative pairs as our evaluation dataset.

\subsection{Experimental results}
% According to the testing protocol of LFW, we give the verification accuracy on $6000$ face pairs.
% %
% Besides different illuminations, occlusions and expressions, cross-pose face is another challenge in face recognition.
% %
% Therefore the Cross-Pose LFW (CPLFW) is proposed which deliberately searches and selects $6,000$ face pairs.
% %
% The testing of AgeDB consists of 5 different year gaps. Each group has ten splits of faces and each split contains $300$ positive examples and $300$ negative examples.
% %
% We choose one of testing subset, AgeDB-30, as our validation dataset.
%
As shown in Table~\ref{result}, the pre-trained teacher model reaches $99.73\%$ on LFW, $92.85\%$ on CPLFW, and~$98.25\%$ on AgeDB-30.
The student model trained by ArcFace reaches $98.75\%$ on LFW, $78.53\%$ on CPLFW, and $93.53\%$ on AgeDB-30.
% 
% Considering that the student network has much less parameters, the ability of extracting features from cross-pose faces decreases.

For comparison with the original triplet loss, we set the fixed margin~$m$ as $0.3$, $0.4$, and $0.5$, which are chosen based on our validation for the best performance of triplet loss.
After applying the proposed triplet distillation to the student model, its verification accuracy is boosted to $99.27\%$ on LFW, $81.28\%$ on CPLFW, and $94.37\%$ on AgeDB-30.
In other words, the accuracy of triplet distillation is consistently higher than the original triplet loss using the fixed margin.

\vspace{-1.2em}
\begin{table}[h]
\caption{Comparison of the proposed Triplet Distillation with triplet loss.}
\label{result}
\vskip 0.15in
\begin{center}
\begin{small}
%\begin{sc}
\begin{tabular}{lccc}
\toprule
Model & LFW & AgeDB-30 & CPLFW\\
\midrule
Teacher  & 99.73\% & 98.25\% & 92.85\% \\
Student  & 98.75\% & 93.53\% & 78.53\% \\
\midrule
Student+Triplet loss \\
margin=0.3  & 99.21\% & 94.08\% & 80.80\% \\
margin=0.4  & 99.23\% & 94.00\% & 81.16\% \\
margin=0.5  & 99.20\% & 93.80\% & 80.38\% \\
\midrule
Student+Ours  & \textbf{99.27\%} & \textbf{94.25\%} & \textbf{81.28\%} \\
\bottomrule
\end{tabular}
%\end{sc}
\end{small}
\end{center}
\vskip -0.1in
\end{table}

%

%
%The experiment results confirm the improvement of triplet loss by our triplet distillation, which maps the similarity between the face representations extracted by the teacher model into these dynamic margins, thus bringing more precise face relationships to the student network.

%

\section{Conclusion}
\label{conc}
We propose triplet distillation for deep face recognition, which takes advantage of knowledge distillation to generate dynamic margins to enhance triplet loss.
The distance obtained by the teacher model reflects similarity information between different identities, which can be regarded as a new type of knowledge.
Compared with the original triplet loss, experiments have proven that our proposed method delivers an encouraging performance improvement.

\bibliography{reference}
\bibliographystyle{icml2019}

\end{document}